# Fusion of Wavelet Coefficients from Visual and Thermal Face Images for Human Face Recognition – A Comparative Study

**Mrinal Kanti Bhowmik**                    mkb_cse@yahoo.co.in
*Lecturer, Department of Computer Science and Engineering*
*Tripura University (A Central University)*
*Suryamaninagar – 799130, Tripura, India*

**Debotosh Bhattacharjee**                    debotosh@indiatimes.com
*Reader, Department of Computer Science and Engineering*
*Jadavpur University*
*Kolkata – 700032, India*

**Mita Nasipuri**                    mitanasipuri@gmail.com
*Professor, Department of Computer Science and Engineering*
*Jadavpur University*
*Kolkata – 700032, India*

**Dipak Kumar Basu**                    dipakkbasu@gmail.com
*Professor, AICTE Emeritus Fellow, Department of Computer Science and Engineering*
*Jadavpur University*
*Kolkata – 700032, India*

**Mahantapas Kundu**                    mkundu@cse.jdvu.ac.in
*Professor, Department of Computer Science and Engineering*
*Jadavpur University*
*Kolkata – 700032, India*

## Abstract

In this paper we present a comparative study on fusion of visual and thermal images using different wavelet transformations. Here, coefficients of discrete wavelet transforms from both visual and thermal images are computed separately and combined. Next, inverse discrete wavelet transformation is taken in order to obtain fused face image. Both Haar and Daubechies (db2) wavelet transforms have been used to compare recognition results. For experiments IRIS Thermal/Visual Face Database was used. Experimental results using Haar and Daubechies wavelets show that the performance of the approach presented here achieves maximum success rate of 100% in many cases.

**Keywords:** Thermal image, Daubechies Wavelet Transform, Haar Wavelet Transform, Fusion, Principal Component Analysis (PCA), Multi-layer Perceptron, Face Recognition.

## 1.  INTRODUCTION

Research activities on human face recognition are growing rapidly and researchers have expanded the application potential of the technology in various domains [1, 2, 3, 4, 5]. It is still a very challenging task to perform efficient face recognition in the outside environments with no control on illumination [6]. Accuracy of the face recognition system degrades quickly when the lighting is dim or when it does not uniformly illuminate the face [7]. Light reflected from human faces also varies depending on the skin color of people from different ethnic groups. Fusion techniques have many advantages over the thermal and visual images to improve the overall recognition accuracy. As a matter of fact fusion of images has already established its importance in case of image analysis, recognition, and classification. Different fusion levels are being used, for example, low-level fusion, high-level fusion etc. Low-level data fusion combines the data from two different images of same pose and produces a new data that contains more details. High-level decision fusion combines the decisions from multiple classification modules [8], [9]. Decision fusion can be accomplished with majority voting, ranked-list combination [10], and the use of Dempster-Shafer theory [11]. Several fusion methods have been attempted in face recognition. Few of them are shown at Table I.

**TABLE 1:** Different Fusion Methods.

| Author | Technique | References |
|--------|-----------|-----------|
| I. Pavlidis | Dual band Fusion System | [6] |
| I. Pavlidis | Near Infrared Fusion Scheme | [12] |
| J. Heo | Data and Decision Fusion | [11] |
| A. Gyaourova | Pixel based fusion using Wavelet | [13] |
| A. Selinger | Appearance based Face Recognition | [14] |
| A. Selinger | Fusion of co-registered visible and LWIR | [14] |
| M. Hanif | Optimized Visual and Thermal Image Fusion | [15] |

One of the fusion technique mentioned above, A. Gyaourova et al [13] tried to implement pixel-based fusion at multiple resolution levels. It allows features with different spatial extend to be fused at the resolution at which they are most salient. In this way, important features appearing at lower resolutions can be preserved in the fusion process. This method contains two major steps: one is fusion of IR and visible images and other one is recognition based on the fused images. Fusion is performed by combining the coefficient of Haar wavelet by decomposition of a pair of IR and visible images having equal size. They used Genetic Algorithms (GAs) to fuse the wavelet coefficients from the two spectra. They use GAs for fusion was based on several factors. First, the search space for the image fusion. Second, the problem at hand appears to have many suboptimal solutions. They shown the effectiveness of the fusion solution by GAs and they perform the recognition using eigenfaces. They used Euclidian distance to recognize the PCA applied by projected data. A. Selinger et al [14] they have compared different algorithms like eigenfaces (PCA), Linear Discriminate Analysis (LDA) and many more for multiple appearance-based face recognition methodology. All the experimental data were captured with a visible CCD array and LWIR microbolometer and used n-fold cross validation technique for the performance. I. Pavlidis et al [6], they propose a novel imaging solution which address the problems like lightning and disguise effect in facial image. They demonstrate a dual-band fusion system in the near infrared which segment human faces much more accurately than traditional visible band face detection systems and they have also represent with theoretical and experimental arguments that the upper band of the near infrared is particularly advantageous for disguise detection purpose. In their face detection system is the only dual band system in the near infrared range. For this method they calls two cameras one for upper-band (within a range of 0.8 - 1.4 μm) and lower-band (within a range of 1.4 - 2.2μm). They generated fused image by using weighted difference method of the co-registered imaging signals from the lower and upper-band cameras because of the abrupt change in the reflectance for the human skin around 1.4 μm the fusion has as a result the intensification of the humans face silhouettes and other exposed skin parts and the diminution of the background. They used Otsu thresholding algorithm for perfect image segmentation. The dual-band near-infrared face detection method could be used in combination with a traditional visible spectrum face recognition method. This approach will provide a much more accurate face segmentation result than a traditional visible band method. G. Bebis et al [16] in this paper they have extended their previous technique which is mentioned at [13]. In this paper, they have introduced another fusion scheme as feature-based fusion which is operates in the eigenspace domain. In this case also they employ a simple and general framework based on Genetic Algorithms (GAs) to find an optimum fusion strategy and for this feature level fusion they use eigenspace domain, which involves combining the eigenfeatures from the visible and IR images. First they compute two eigenspaces, of visible IR face images. They generate the fused image by applying the GAs to the eigenfeatures from the IR and visible eigenspace. In another scheme, M. Hanif et al [15] proposed data fusion of visual and thermal images using Gabor filtering technique. It has been found that by using the proposed fusion technique Gabor filter can recognize face even with variable expressions and light intensities, but not in extreme condition. They have designed the data fusion technique in the spatial and DWT domain. For the spatial domain they have used a weight to generate the fusion image. But in case of DWT image fusion, they first decompose both thermal and visual image up to level-n and then choose the absolute maximum at every pixel for both the thermal and visual image to generate the fusion image and this work is implemented using equinox face database. Jingu Heo et al [11] describe two types of fusion method as Data fusion and Decision Fusion. In data fusion, it produces an illumination-invariant face images by integrating the visual and thermal images. A weighted average technique is used for data fusion. In the data fusion they first detect the eyeglasses and remove the eyeglasses by a fitting an eclipse and then merge it with visual image. They used n non-iterative eclipse fitting algorithm for it. They have implemented Decision fusion with highest matching score (Fh) and Decision fusion with average matching score (Fa). Comparison results on three fusion-based face recognition techniques showed that fusion-based face recognition techniques outperformed individual visual and thermal face recognizers under illumination variations and facial expressions. From them Decision fusion with average matching score consistently demonstrated superior recognition accuracies as per their results. The organization of the rest of this paper is as follows: in section 2, the overview of the system is discussed, in section 3 experimental results and discussions are given. Finally, section 4 concludes this work.

## 2. SYSTEM OVERVIEW

The block diagram of the system is given in Figure 1. All the processing steps used in this work are shown in the block diagram. In the first step, decomposition of both the thermal and visual images up to level five has been done using Haar and Daubechies wavelet transformations. Then fused images are generated using respective decomposed visual and thermal images of two different wavelets. These transformed images separated into two groups namely, training set and testing set are fed into a multilayer perceptron for classification. Before, feeding them into the network, those are passed through Principal Component Analysis (PCA) for dimensionality reduction.

### 2A. Wavelet Transforms and Fusion

Haar wavelet is recognized as the first known wavelet. It is same as Daubechies wavelet (db1). The Haar wavelet is proposed by Alfred Haar [17]. Haar wavelet is a certain sequence of function.

The Haar wavelet's mother wavelet function $\psi(t)$ can be described as

$$\psi(t) = \begin{cases} 1 & 0 \le t < 1/2. \\ -1 & 1/2 \le t < 1. \\ 0 & otherwise. \end{cases} \quad (1)$$

and its scaling function $\phi(t)$ can be described as

$$\phi(t) = \begin{cases} 1 & 0 \le t < 1, \\ 0 & otherwise. \end{cases} \quad (2)$$

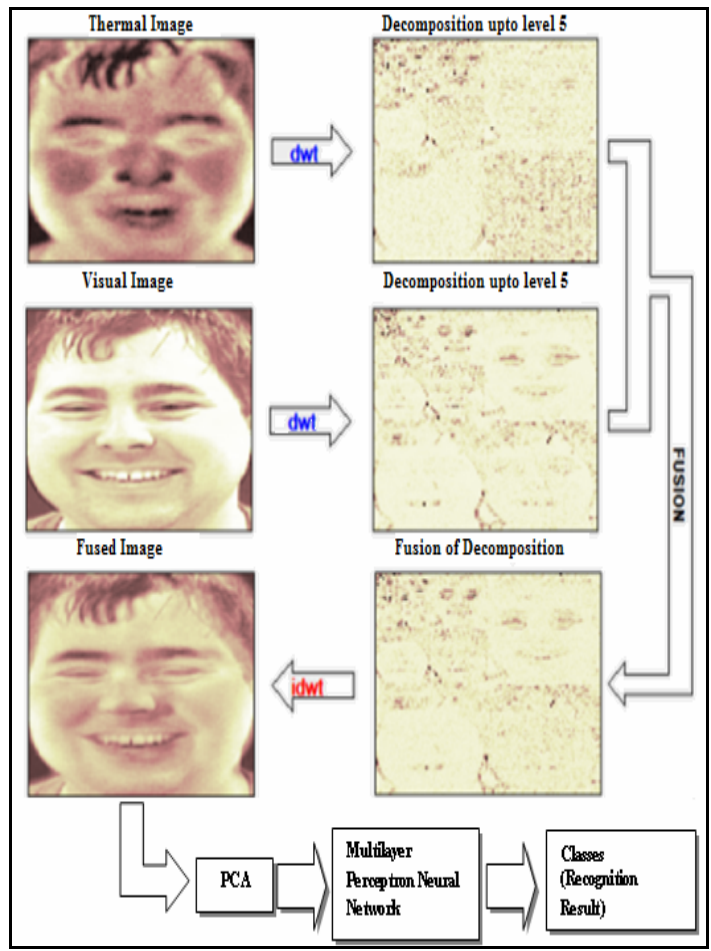

**FIGURE 1:** System Overview.

It is an orthogonal and bi-orthogonal wavelet. The technical disadvantage of Haar wavelet is that it is not continuous, and therefore not differentiable. In functional analysis, the Haar systems denotes the set of Haar wavelets

$$\{t \mapsto \psi_{n,k}(t) = \psi(2^n t - k); n \in \mathbb{N}, 0 \le k < 2^n\}$$ 

(3)

The Haar system (with the natural ordering) is further a Schauder basis for the space Lp[0, 1] for $1 \le p < +\infty$. This basis is unconditional for p > 1 [18], [19].

The Haar transform is the simplest of the wavelet transforms. This transform cross-multiplies a function against the Haar wavelet with various shifts and stretches. The Haar transform is derived from the Haar matrix. An example of 3x3 Haar matrixes is shown below:

$$H_1 = \frac{1}{\sqrt{4}}\begin{bmatrix} 1 & 1 & 1 & 1 \\ 1 & 1 & -1 & -1 \\ \sqrt{2} & -\sqrt{2} & 0 & 0 \end{bmatrix}$$ 

(4)

The Haar transform can be thought of as a sampling process in which rows of the transform matrix act as samples of finer and finer resolution.

Daubechies wavelets are a family of orthogonal wavelets defining a discrete wavelet transform and characterized by a maximal number of vanishing moments for some given support. In two dimensions, the difference of information between the approximations $A_{2^{j+1}}f$ and $A_{2^{j}}f$ is therefore characterized by the sequences of inner products

$$D_{2^j}^1 f = \left( \langle f(x,y), \Psi_{2^j}^1 (x - 2^{-j}n, y - 2^{-j}m) \rangle \right)_{(n,m)\epsilon Z^2} \tag{5}$$

$$D_{2^j}^2 f = \left( \langle f(x,y), \psi_{2^j}^2 (x - 2^{-j}n, y - 2^{-j}m) \rangle \right)_{(n,m)\epsilon Z^2} \tag{6}$$

$$D_{2^j}^3 f = \left( \langle f(x,y), \psi_{2^j}^3 (x - 2^{-j}n, y - 2^{-j}m) \rangle \right)_{(n,m)\epsilon Z^2} \tag{7}$$

Each of these sequences of inner products can be considered as an image. In Figure 2(a) showing level 2 decomposition, $H_2$ gives the vertical higher frequencies (horizontal edges), $V_2$ gives the horizontal higher frequencies (vertical edges), and $D_2$ gives higher frequencies in both directions (corners). Let us suppose that initially we have an image $A_1\ f$ measured at the resolution 1. For any $J > 0$, this discrete image can be decomposed between the resolution 1 and $2^{-J}$, and completely represented by the $3J + 1$ discrete images

$$\left( A_{2^{-J}}f, \left(D_{2^j}^1 f\right)_{-J \le j \le -1}, \left(D_{2^j}^2 f\right)_{-J \le j \le -1}, \left(D_{2^j}^3 f\right)_{-J \le j \le -1} \right) \tag{8}$$

This set of images is called an orthogonal wavelet representation in two dimensions [20]. The image $A_{2^{-j}}f$ is a coarse approximation, and the images $H_2$ give the image details for different orientations and resolutions. If the original image has $N^2$ pixels, each image $A_{2^j}\ f$, $H_2$, $V_2$, $D_2$ has $2^j * N^2$ pixels (j < 0). The total number of pixels of an orthogonal wavelet represents is therefore equal to $N^2$. It does not increase the volume of data. This is due to the orthogonality of the representation. Such decompositions were done through Matlab and shown in equations 9 and 10.

$$[cA, cH, cV, cD] = dwt2 \ (X, \ 'wname') \tag{9}$$

$$[cA, cH, cV, cD] = dwt2 \ (X, \ Lo\_D, \ Hi\_D) \tag{10}$$

Equation (9), 'wname' is the name of the wavelet used for decomposition. We used 'haar' and 'db2' as wavelet. Equation (10) Lo_D (decomposition low-pass filter) and Hi_D (decomposition high-pass filter) wavelet decomposition filters [21], [22], [23], [35], [39].

The more the decomposition scheme is being repeated, the more the approximation of images concentrates in the low frequency energy. During the decomposition process it actually down-sampling the rows and columns of an image. Firstly it down-samples the rows (keep one out of two) and then down-samples the columns (keeps one out of two). The downsampling process for both the images is same. After decomposing the thermal and visual image we have to generate the decomposed fused image by providing fusion methods. Then, at each level select the absolute 'maximum' at every pixel from both the images for approximate co-efficient (cA) and absolute 'minimum' at every pixel from both the images for three details co-efficient (cH, cV, cD). Let us take an example, if T is the thermal and V is the visual image with same pose and illumination than

$$\text{'max': } D = abs(T) \ge abs(V); \ C = T(D) + V(\sim D) \tag{11}$$

$$\text{'min': } D = abs(T) \le abs(V); \ C = T(D) + V(\sim D) \tag{12}$$

To generate the fused image (C) of co-efficient it will add that value which is deducted from the visual image during the calculation of absolute value (D) of thermal (T) and visual (V) image for all the coefficients using fusion method shown at equation (11) & (12).

Consequently the reconstruction process is performed using inverse of wavelet transformations to generate synthesized fused image.

$$X = idwt2 \ (cA, cH, cV, cD, 'wname') \qquad (13)$$

$$X = idwt2 \ (cA, cH, cV, cD, Lo\_R, Hi\_R) \qquad (14)$$

IDWT uses the wavelet 'wname' to compute the single-level reconstruction of an Image X, based on approximation matrix (cA) and detailed matrices cH, cV and cD (horizontal, vertical and diagonal respectively). By the equation no (14), we can reconstruct the image using filters Lo_R (reconstruct low-pass) and Hi_R (reconstruct high-pass) and 'haar' and 'db2' as the time of reconstruction of an image [33].

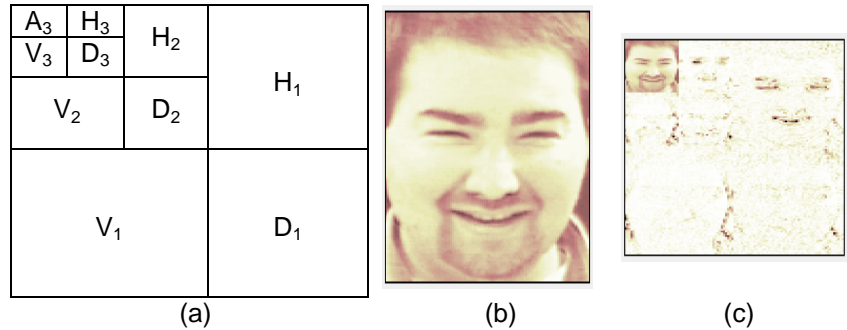

**FIGURE 2:** (a) Shown one of the third level decomposition of level 5 (b) Original image used in the decomposition (c) Orthogonal wavelet representation of a sample image.

The decomposition process can be iterated, with successive approximations being decomposed in turn, so that one image is broken down into many lower resolution components. This is called the wavelet decomposition tree. In this work, decomposition was done up to level five using Haar and Daubechies (db2) wavelet transformations, as shown in Figure 3. Average of corresponding transform coefficients from visual and thermal images gives the matrix of fused coefficients, which when transformed into the image in spatial domain by inverse wavelet transform produces fused face image. These fused images thus found are passed through PCA for dimensionality reduction, which is described next.

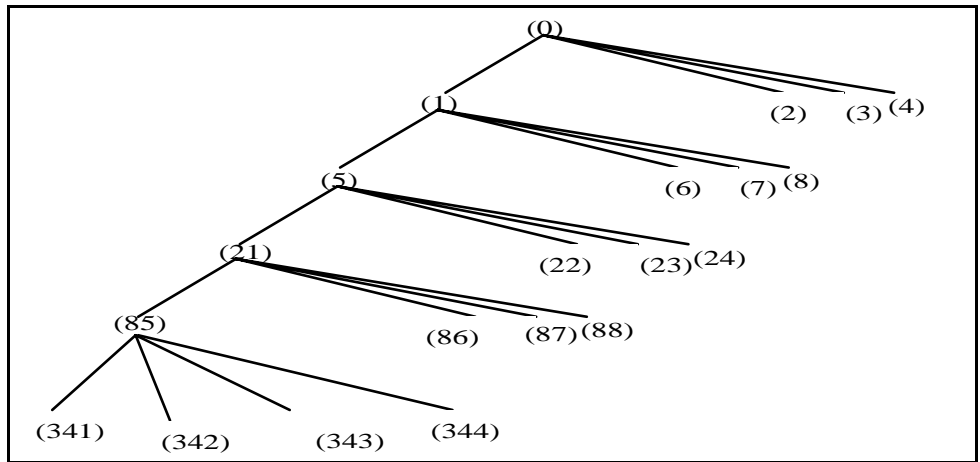

**FIGURE 3:** Wavelet Decomposition Tree.

### 2B. Dimensionality Reduction

Principal component analysis (PCA) is based on the second-order statistics of the input image, which tries to attain an optimal representation that minimizes the reconstruction error in a least-squares sense. Eigenvectors of the covariance matrix of the face images constitute the eigenfaces. The dimensionality of the face feature space is reduced by selecting only the eigenvectors possessing significantly large eigenvalues. Once the new face space is constructed, when a test image arrives, it is projected onto this face space to yield the feature vector—the representation coefficients in the constructed face space. The classifier decides for the identity of the individual, according to a similarity score between the test image's feature vector and the PCA feature vectors of the individuals in the database [27], [30], [36], [37], [34].

### 2C. Artificial Neural Network Using Backpropagation With Momentum

Neural networks, with their remarkable ability to derive meaning from complicated or imprecise data, can be used to extract patterns and detect trends that are too complex to be noticed by either humans or other computer techniques. A trained neural network can be thought of as an "expert" in the category of information it has been given to analyze. The Back propagation learning algorithm is one of the most historical developments in Neural Networks. It has reawakened the scientific and engineering community to the modeling and processing of many quantitative phenomena. This learning algorithm is applied to multilayer feed forward networks consisting of processing elements with continuous differentiable activation functions. Such networks associated with the back propagation learning algorithm are also called back propagation networks [24], [25], [26], [27], [28], [29], [38].

## 3. EXPERIMENTAL RESULTS AND DISCUSSIONS

This work has been simulated using MATLAB 7 in a machine of the configuration 2.13 GHz Intel Xeon Quad Core Processor and 16384.00 MB of Physical Memory. We analyze the performance of our algorithm using the IRIS thermal / visual face database. In this database, all the thermal and visible unregistered face images are taken under variable illuminations, expressions, and poses. The actual size of the images is 320 x 240 pixels (for both visual and thermal). Total 30 classes are present in that database [31].  Some thermal and visual images and their corresponding fused images for Haar and Daubechies (db2) wavelets are shown in Figure 4 and Figure 5 respectively.

To compare results of Haar and Daubechies (db2) wavelets, fusion of visual and thermal images were done separately.  For that purpose 200 thermal and 200 visual images were considered. We have increased the size of testing image for both the wavelet. Firstly we have tested our system using 100 fused images using two different wavelet and we increase 10 images per class for 10 different classes i.e. total 100 image. The database is categorized into 10 classes based on changes in illumination and expressions. The classes with illuminations and expression changes are class–1, class–2, class–3, class–4, class–6, class–7, and class–9, whereas class–5, class–8 and class–10 are with changes in expressions only. Both the techniques were applied and experimental results thus obtained are shown in Figure 6.

After increasing the number of testing images the recognition rates are increased for both the wavelets. Previously we have average recognition rate is 85% for Haar wavelet and 90.019% for Daubechies wavelet when the number of testing image is 100 (i.e. 10 image per class). Now we have average recognition rate is 87% for Haar and 91.5% for Daubechies wavelet after increase the number of testing image 100 to 200 (i.e. 20 image per class).

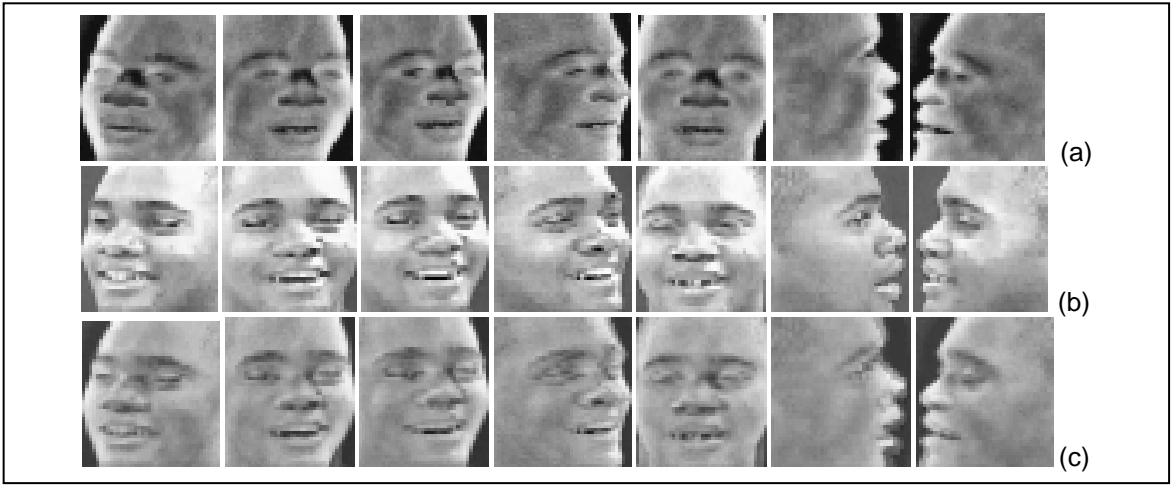

**FIGURE 4:** Sample images of IRIS database (a) Thermal images (b) Visual images (c) Corresponding Fused images using Haar Wavelet transform.

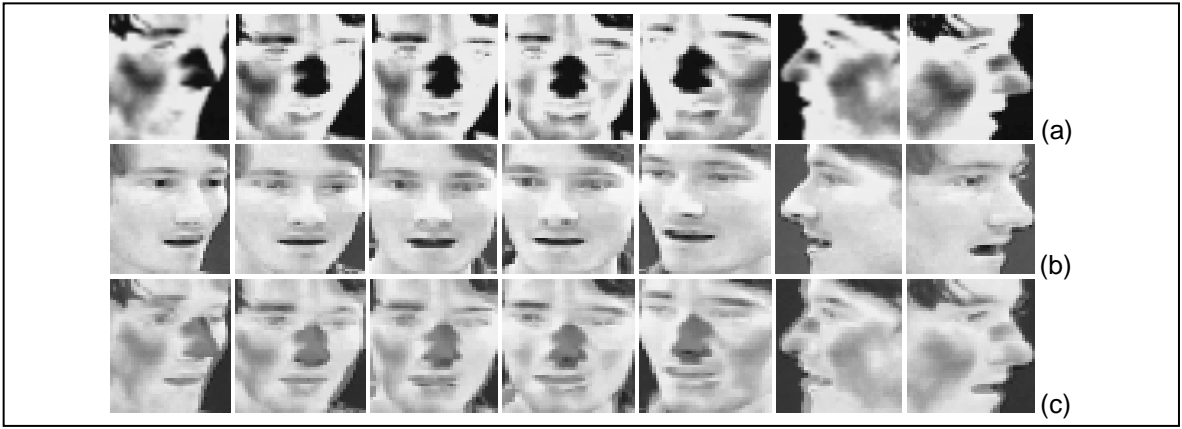

**FIGURE 5:** Sample images of IRIS database (a) Thermal images (b) Visual images (c) Corresponding Fused images using Daubechies Wavelet transform.

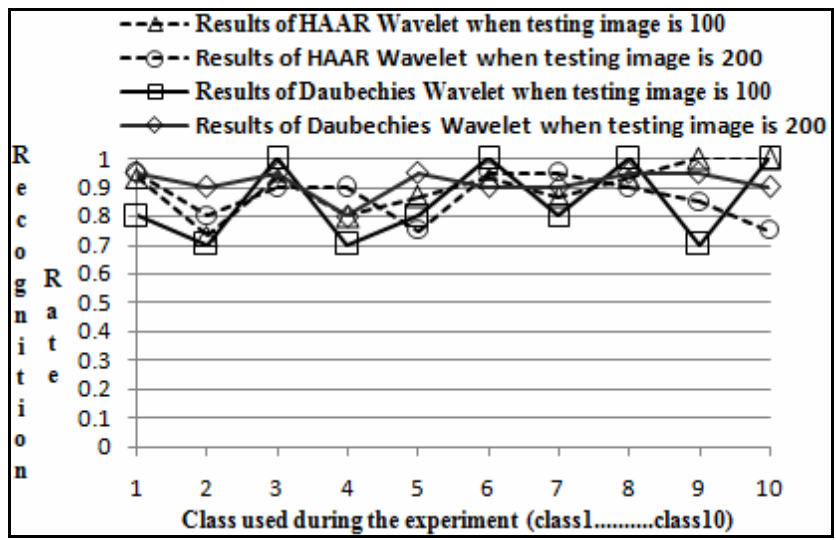

**FIGURE 6:** Recognition rate of Haar and Daubechies Wavelet.

In the Figure 6, all the recognition rates for Haar as well as Daubechies wavelet transformations have been shown. In this figure "dotted" line indicates results using Haar wavelet (with round and triangle shape marker) and "solid" line indicates results using Daubechies Wavelet (with box and diamond shape marker). From this figure, it can be easily inferred that overall results using "Daubechies" (db2) wavelet produces better result than "Haar" wavelet.

Final results have been computed as average of all the categories and given along with results of other recent methods in Table II. Here, we find results using Daubechies wavelets (db2) shows comparable results.

**TABLE 2:** Comparison between different Fusion Techniques.

| Image Fusion Technique | | Recognition Rate | | | |
|---|---|---|---|---|---|
| Present method | Haar | 85% | When testing image 100 | 87% | When testing image 200 |
| | Daubechies | 90.019% | | 91.5% | |
| Simple Spatial Fusion[15] | | 91.00% | | | |
| Fusion of Thermal and Visual [32] | | 90.00% | | | |
| Abs max selection in DWT[15] | | 90.31% | | | |
| Window based absolute maximum selection [15] | | 90.31% | | | |
| Fusion of Visual and LWIR + PCA [6] | | 87.87% | | | |

# 4. CONSLUSION

In this paper, a comparative study on two different wavelet transformations, namely, Haar and Daubechies (db2) for image fusion has been presented. After completion of fusion, images were projected into an eigenspace. Those projected fused eigenfaces are classified using a Multilayer Perceptron. Eigenspace is constituted by the images belong to the training set of the classifier, which is a multilayer perceptron. The efficiency of the scheme has been demonstrated on IRIS system has achieved a maximum recognition rate of 100% using both the wavelets i.e. Haar and Daubechies. But on an average 85% and 90.019% (using 100 testing images) and 87% and 91.5% (using 200 testing images) recognition rates were achieved for Haar wavelet and Daubechies wavelet respectively.

# 5. ACKNOWLEDGMENT

Mr. M. K. Bhowmik is thankful to the project entitled "Development of Techniques for Human Face Based Online Authentication System Phase-I" sponsored by Department of Information Technology under the Ministry of Communications and Information Technology, New Delhi-110003, Government of India Vide No. 12(14)/08-ESD, Dated 27/01/2009 at the Department of Computer Science & Engineering, Tripura University-799130, Tripura (West), India for providing the necessary infrastructural facilities for carrying out this work. The first author would also like to thank Prof. Barin Kumar De, H.O.D. Physics and Dean of Science of Tripura University, for his kind support to carry out this research work.

# 6. REFERENCES

1.  A. Pentland and T. Choudhury. "Face Recognition for Smart Environments". IEEE Computer. 33(2):50-55, 2000

2.  P. Phillips et al., The Feret Database and Evaluation Procedure for Face Recognition Algorithms. Image and Vision Computing. May 1998. pp. 295-306

3.  L. Wiskott et al., "Face Recognition by Elastic Bunch Graph Matching". Trans. IEEE Pattern Analysis and Machine Intelligence. 19(7):775-779. 1997

4.  B. Moghaddam and A. Pentland. "Probabilistic Visual Recognition for Object Recognition". Trans. IEEE Pattern Analysis and Machine Intelligence. 19(7):696-710. 1997

5.  P. Penev and J. Atick. "Local Feature Analysis: A General Statistical Theory for Object Representation". Network: Computation in Neural Systems. Mar. 1996, pp. 477-500

6.  I. Pavlidis and P. Symosek. "The Imaging Issue in an Automatic Face/Disguise Detecting System". Proceedings of the IEEE Workshop on Computer Vision Beyond the Visible Spectrum: Methods and Applications (CVBVS 2000)

7.  Y. Adini, Y. Moses, and S. Ullman. "Face recognition: The problem of compensating for changes in illumination direction". IEEE Trans. On Pattern Analysis and Machine Intelligence. Vol. 19, No. 7. pp.721-732, 1997

8.  A. Ross and A. Jain. "Information fusion in biometrics". Pattern Recognition Letters. Vol. 24, No. 13. pp.2115-2125. 2003

9.  T. K. Ho, J. J. Hull, and S. N. Srihari. "Decision combination in multiple classifier systems". IEEE Trans. on Pattern Analysis and Machine Intelligence. Vol. 16. No. 1. pp.66-75. 1994

10. C. Sanderson and K. K. Paliwal. "Information fusion and person verification using speech and face information". IDIAP Research Report 02-33. Martigny. Switzerland. 2002

11. J. Heo, S. G. Kong, B. R. Abidi and M. A. Abidi. "Fusion of Visual and Thermal Signatures with Eyeglass Removal for Robust Face Recognition". Conference on Computer Vision and Pattern Recognition Workshop (CVPRW'04) Vol. 8. pp. 122-122. USA. 2004

12. I. Pavlidis, P. Symosek, B. Fritz. "A Near Infrared Fusion Scheme for Automatic Detection of Humans". U.S. Patent Pending. H16-25929 US. filed in September 3, 1999

13. A. Gyaourova, G. Bebis and I. Pavlidis. "Fusion of infrared and visible images for face recognition". Lecture Notes in Computer Science. 2004.Springer. Berlin

14. A. Selinger, D. Socolinsky. "Appearance-based facial recognition using visible and thermal imagery: a comparative study". Technical Report 02-01. Equinox Corporation. 2002

15. M. Hanif and U. Ali. "Optimized Visual and Thermal Image Fusion for Efficient Face Recognition". IEEE Conference on Information Fusion. 2006

16. G. Bebis, A. Gyaourova, S. Singh and I. Pavlidis. "Face Recognition by Fusing Thermal Infrared and Visible Imagery". Image and Vision Computing. Vol. 24, Issue 7. July 2006. pp. 727-742

17. A. Haar. "Zur Theorie der orthogonalen Funktionensysteme (German)"

18. http://eom.springer.de/O/o070308.htm

19. W. G. Gilbert, S. Xiaoping. "Wavelets and Other Orthogonal Systems". 2001

20. T. Paul. "Affine coherent statesand the radial Schrodinger equation. Radial harmonic oscillator and hydrogen atom". Preprint

21. I. Daubechies. Ten lectures on wavelets, CBMS-NSF conference series in applied mathematics. SIAM Ed. 1992

22. S. G. Mallat. "A theory for multiresolution signal decomposition: The wavelet representation". IEEE Trans. Pattern Anal. Machine Intell., vol.11. pp. 674-693. July 1989

23. Y. Meyer. Wavelets and operators. Cambridge Univ. Press. 1993

24. M. K. Bhowmik, D. Bhattacharjee, M. Nasipuri, D.K. Basu and M. Kundu. "Classification of Polar-Thermal Eigenfaces using Multilayer Perceptron for Human Face Recognition". in Proceedings of the 3rd IEEE Conference on Industrial and Information Systems (ICIIS-2008). IIT Kharagpur. India. Dec. 8-10. 2008. Page no: 118

25. D. Bhattacharjee, M. K. Bhowmik, M. Nasipuri, D. K. Basu and M. Kundu. "Classification of Fused Face Images using Multilayer Perceptron Neural Network". Proceedings of International Conference on Rough sets. Fuzzy sets and Soft Computing. November 5–7, 2009. organized by Deptt. of Mathematics. Tripura University. pp. 289 – 300

26. M. K. Bhowmik, D. Bhattacharjee, M. Nasipuri, D.K. Basu and M. Kundu. "Classification of Log Polar Visual Eigenfaces using Multilayer Perceptron". Proceedings of The 2nd International Conference on Soft computing (ICSC-2008). IET. Alwar. Rajasthan. India. Nov. 8–10. 2008. pp:107-123

27. M. K. Bhowmik, D. Bhattacharjee, M. Nasipuri, D. K. Basu and M. Kundu. "Human Face Recognition using Line Features". proceedings of National Seminar on Recent Advances on Information Technology (RAIT-2009) Feb. 6-7,2009. Indian School of Mines University. Dhanbad. pp: 385

28. M. K. Bhowmik. "Artificial Neural Network as a Soft Computing Tool – A case study". In Proceedings of National Seminar on Fuzzy Math. & its application. Tripura University. Nov. 25 – 26, 2006. pp: 31 – 46

29. M. K. Bhowmik, D. Bhattacharjee and M. Nasipuri. "Topological Change in Artificial Neural Network for Human Face Recognition". Proceedings of National Seminar on Recent Development in Mathematics and its Application. Tripura University. Nov. 14 – 15, 2008. pp: 43 – 49

30. H. K. Ekenel and B. Sankur. "Multiresolution face recognition". Image and Vision Computing. 2005. Vol. 23. pp. 469 – 477

31. http://www.cse.ohio-state.edu/otcbvs-bench/Data/02/download.html

32. A. Singh, G. B. Gyaourva and I. Pavlidis. "Infrared and Visible Image Fusion for Face Recognition". Proc. SPIE. vol.5404. pp.585-596. Aug.2004

33. V. S. Petrovic and C. S. Xydeas. "Gradient-Based Multiresolution Image Fusion". IEEE Trans. On image Processing. Vol. 13. No. 2. Feb. 2004.

34. A. Rahim & J. V. Charangatt. "Face Hallucination using Eigen Transformation in Transform Domain". International Journal of Image Processing (IJIP) Vol. 3. Issue 6, pp. 265-282.

35. J. Shen and G. Strang. "Applied and Computational Harmonic Analysis". 5(3). "Asymptotics of Daubechies Filters, Scaling Functions and Wavelets"

36. M. K. Bhowmik, D. Bhattacharjee, M. Nasipuri, D. K. Basu and M. Kundu. "Classification of fused images using Radial basis function Neural Network for Human Face Recognition".

Proceedings of The World congress on Nature and Biologically Inspired Computing (NaBIC-09). Coimbatore, India. published by IEEE Computer Society. Dec. 9-11, 2009. pp. 19-24

37. M. K. Bhowmik, D. Bhattacharjee, M. Nasipuri, D.K. Basu and M. Kundu. "Image Pixel Fusion for Human Face Recognition". appeared in International Journal of Recent Trends in Engineering [ISSN 1797-9617]. published by Academy Publishers. Finland. Vol. 2. No. 1, pp. 258–262. Nov. 2009

38. D. Bhattacharjee, M. K. Bhowmik, M. Nasipuri, D. K. Basu & M. Kundu. "A Parallel Framework for Multilayer Perceptron for Human Face Recognition". International Journal of Computer Science and Security (IJCSS), Vol. 3. Issue 6, pp. 491-507.

39. Sathesh & Samuel Manoharan. "A Dual Tree Complex Wavelet Transform Construction and its Application to Image Denoising". International Journal of Image Processing (IJIP) Vol. 3. Issue 6, pp. 293-300.